# Sparsity Constrained Graph Regularized NMF for Spectral Unmixing of Hyperspectral Data

*Roozbeh Rajabi, Hassan Ghassemian*

Abstract

Hyperspectral images contain mixed pixels due to low spatial resolution of hyperspectral sensors. Mixed pixels are pixels containing more than one distinct material called endmembers. The presence percentages of endmembers in mixed pixels are called abundance fractions. Spectral unmixing problem refers to decomposing these pixels into a set of endmembers and abundance fractions. Due to nonnegativity constraint on abundance fractions, nonnegative matrix factorization methods (NMF) have been widely used for solving spectral unmixing problem. In this paper we have used graph regularized NMF (GNMF) method combined with sparseness constraint to decompose mixed pixels in hyperspectral imagery. This method preserves the geometrical structure of data while representing it in low dimensional space. Adaptive regularization parameter based on temperature schedule in simulated annealing method also has been used in this paper for the sparseness term. Proposed algorithm is applied on synthetic and real datasets. Synthetic data is generated based on endmembers from USGS spectral library. AVIRIS Cuprite dataset is used as real dataset for evaluation of proposed method. Results are quantified based on spectral angle distance (SAD) and abundance angle distance (AAD) measures. Results in comparison with other methods show that the proposed method can unmix data more effectively. Specifically for the Cuprite dataset, performance of the proposed method is approximately 10% better than the VCA and Sparse NMF in terms of root mean square of SAD.

1. Introduction

Mixed pixels appear in hyperspectral images due to low spatial resolution of hyperspectral sensors (Keshava 2003). Fig. 1 illustrates the basic concept of mixed pixels in hyperspectral images (Rajabi and Ghassemian 2011). Pure pixel refers to a pixel that is composed of only one distinct material and mixed pixel refers to a pixel containing more than one distinct material. Spectral unmixing problem has many applications in hyperspectral data analysis, for example it can be used to classify the hyperspectral data at a finer spatial resolution (Villa et al. 2011), hyperspectral and multispectral image fusion (Bendoumi and Mingyi 2013).

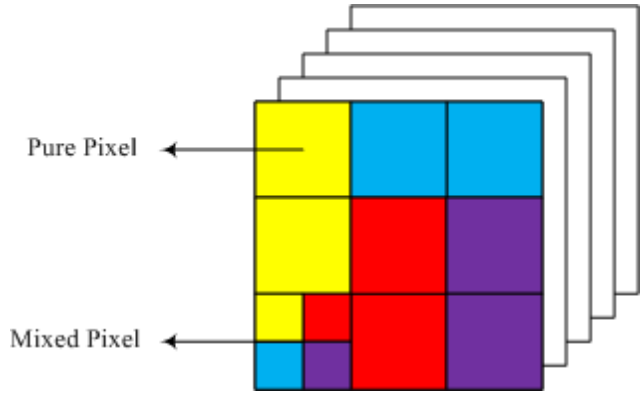

Fig. 1. Pure and mixed pixels concept in hyperspectral imagery

Spectral unmixing algorithms decompose a mixed pixel into a set of endmembers and abundance fraction maps (Sanjeevi and Barnsley 2000). Fig. 2 shows a toy example that demonstrates spectral unmixing process. Endmembers are the spectral signatures that are present in the scene and abundance fractions are the percentage of the endmembers in each mixed pixel. Spectral unmixing problem is subject to two constraints on abundance fractions. Firstly, abundance fraction values for each endmember should be nonnegative. Secondly, endmembers within each mixed pixel should cover the pixel completely i.e. sum of abundance fraction values for each pixel should be equal to one.

To mathematically model the spectral unmixing problem, linear or nonlinear models can be considered. In linear mixing model, spectral signature of each pixel is models using a linear combination of true signatures that are present in the scene (Remon et al. 2013). On the other hand, nonlinear models assume nonlinear interactions between different materials in the pixel (Chen et al. 2013). Although nonlinear models are more precise than linear model, but in practice linear model is easy to implement and is accurate for many scenarios. In this paper we used linear mixing model (LMM) to solve spectral unmixing problem.

There are many methods presented in the literature for spectral unmixing (Ma et al. 2014). These methods can be categorized into two main categories. Methods that assume there are pure pixels for each endmember in the scene and methods that works on highly mixed data i.e. all the pixels in the scene are mixed. Vertex component analysis (VCA) (Nascimento and Dias 2005) is one of the well-established methods in the first category. This method also can be categorized as a geometrical method. Blind source separation methods like ICA are also applied to the spectral unmixing problem but as these algorithms need independence assumption they have to be modified to give promising results (Li and Yin 2013).

Nonnegative matrix factorization (NMF) methods have attracted much attention due to nonnegativity constraint on abundance fractions in spectral unmixing problem (Pauca et al. 2006). Several algorithms proposed in the literature based on NMF (Miao and Qi 2007; Yang et al. 2011; Alizadeh and Ghassemian 2012). In (Cai et al. 2011), the authors proposed using graph regularization term for data representation. Based on this idea, we proposed using graph regularized NMF (GNMF) method for hyperspectral unmixing in (Rajabi and Ghassemian 2011). To improve the results of the algorithm we have also added sparseness constraint to GNMF method in (Rajabi and Ghassemian 2013). In the current study, the idea of using adaptive regularization parameter for sparseness constraint is used and examined for hyperspectral unmixing and more experiments have been done to evaluate the proposed method. To quantify the results, root mean square of spectral angle distance (SAD) and abundance angle distance (AAD) are used.

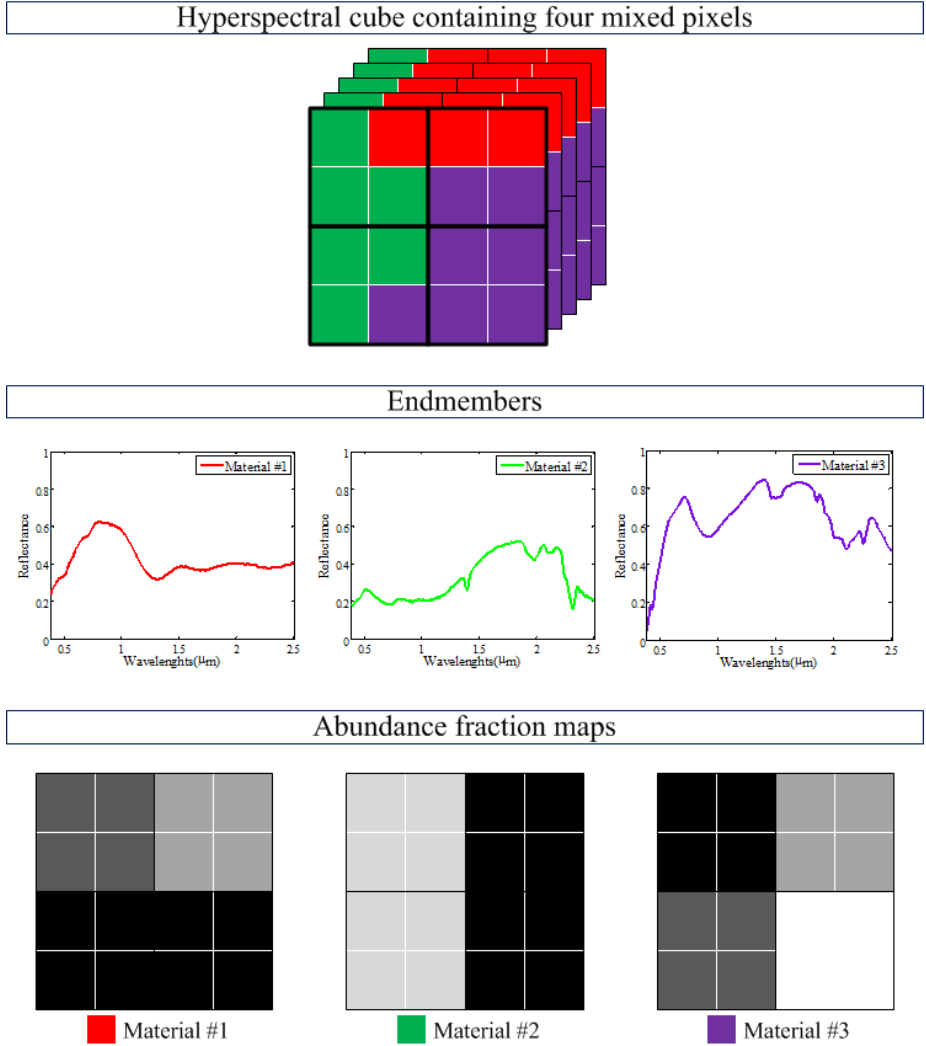

Fig. 2. Toy example explaining spectral unmixing concept

The rest of the paper has been organized as follows. In section II, methodology of the algorithm has been described. In section III, the proposed algorithm has been applied on synthetic and real data and results are presented and compared using quantitative measures. Section IV concludes the paper.

2. Methodology

In this section, firstly mathematical representation of linear mixing model has been described. Then NMF, sparse NMF, GNMF and sparse GNMF are described respectively.

2.1. Linear mixture model (LMM)

There are two models of spectral mixture that can be used for hyperspectral imagery: linear and nonlinear models. In this paper linear mixing model has been used. The mathematical representation of this model is given in the following equation. Table 1 summarizes the variables in the linear mixing model.

$$X = AS + E \tag{1}$$

Table 1. Variables in linear mixture model

| Variable | Description | Dimension |
|---|---|---|
| X | Observed data | L by N |
| A | Endmember spectral signatures | L by P |
| S | Abundance fractions | P by N |
| E | Measurement and sensor noise | L by N |
| N | Total number of pixels | - |
| L | Number of spectral bands | - |
| P | Number of endmembers | - |

LMM is subject to the following constraints that are already discussed in the introduction section: Abundance nonnegativity constraint (ANC) and Abundance sum to one constraint (ASC). These constraints are mathematically expressed for each pixel in the following equations:

$$\text{for } p = 1 \text{ to } P; \; a_p \geq 0 \tag{2}$$

$$\sum_{p=1}^{P} a_p = 1 \tag{3}$$

2.2. Nonnegative matrix factorization (NMF)

Nonnegative matrix factorization (NMF) method is an efficient method for decomposing multivariate data. The mathematical representation of NMF problem is stated in Eq. (4).

$$\mathbf{X} \approx \mathbf{AS} \tag{4}$$

To solve the NMF problem, two different cost functions based on Euclidean and Kullback-Liebler distance can be considered. In this paper the Euclidean distance is used as defined in Eq. (5).

$$O_{NMF} = \|\mathbf{X} - \mathbf{AS}\|_F^2 \tag{5}$$

Multiplicative update rules can be used to optimize the cost function and obtain $\mathbf{A}$ and $\mathbf{S}$ matrices (Pauca et al. 2006).

2.3. Sparse NMF

Since NMF method does not lead to a unique answer, it usually should be utilized with other constraints. One of the most common constraints that can be used for spectral unmixing is sparseness constraint. The cost function of this method is given in Eq. (6) (Qian et al. 2011).

$$O_{SparseNMF} = \|\mathbf{X} - \mathbf{AS}\|_F^2 + \lambda \|\mathbf{S}\|_{1/2} \tag{6}$$

where $\lambda$ is the regularization parameter and determine the impact of the sparseness constraint in the solution.

2.4. Graph regularized NMF (GNMF)

Graph regularized (GNMF) was originally proposed as a feature extraction and dimension reduction technique in (Cai et al. 2011). GNMF preserves the graph structure in the process of decomposition. To model the geometrical structure of the data, an affinity graph is used and new regularization term based on the weight matrix ($\mathbf{W}$) of this graph is added to the cost function as given in Eq. (7).

$$O_{Sparse\ GNMF} = \|\mathbf{X} - \mathbf{AS}\|_F^2 + \mu \mathrm{Tr}(\mathbf{SLS^T}) \tag{7}$$

where $\mathrm{Tr}(.)$ denotes the trace of the matrix and $\mathbf{L} = \mathbf{D} - \mathbf{W}$, $\mathbf{D}_{jj} = \sum_{l} \mathbf{W}_{jl}$. In this cost function $\mu$ is the parameter that controls the impact of graph regularization.

To generate the weight matrix, KNN neighborhood system is used and the weight of edges on the graph is defined using heat kernel function in Eq. (8).

$$\mathbf{W}_{jl} = e^{-\frac{\|\mathbf{x}_j - \mathbf{x}_l\|^2}{\sigma}} \tag{8}$$

where $\mathbf{x}_j$ and $\mathbf{x}_l$ are neighbors and $\sigma$ is the parameter that controls the similarity measure between two points.

## 2.5. Sparse GNMF

In this paper we proposed using sparsity constrained GNMF for the purpose of spectral unmixing. In this method the cost function given in Eq. (9) is used to solve the NMF problem.

$$O_{Sparse\ GNMF} = \|\mathbf{X} - \mathbf{AS}\|_F^2 + \lambda \|\mathbf{S}\|_{1/2} + \mu \mathrm{Tr}(\mathbf{SLS^T}) \tag{9}$$

We used the following equation to update the regularization parameter in each iteration of optimization process.

$$\lambda = \lambda_0 e^{-\frac{t}{\tau}} \tag{10}$$

where $t$ is the iteration number, $\lambda_0$ and $\tau$ are the parameters that controls the impact of sparsity constraint. This equation is motivated by temperature function in the simulated annealing (SA) and successfully applied to blind source separation problems (A Cichocki and Zdunek 2006). As expressed in (A. Cichocki et al. 2006) $\lambda_0$ and $\tau$ should be selected so that $\lambda$ to be in the range of 0.01~0.5. The exact values of these parameters are given in section 3 of the paper.

The cost function of the proposed method is not convex in both $\mathbf{A}$ and $\mathbf{S}$ together, but iterative multiplicative rules can be calculated to minimize the cost function by differentiating it with respect to $\mathbf{A}$ and $\mathbf{S}$ separately. The multiplicative update rules for $\mathbf{A}$ and $\mathbf{S}$ are given in Eq. (11) and (12) respectively.

$$\mathbf{A} = \mathbf{A} .* \frac{\mathbf{XS}^T}{\mathbf{ASS}^T} \tag{11}$$

$$\mathbf{S} = \mathbf{S} .* \frac{\mathbf{A}^T \mathbf{X} + \mu \mathbf{SW}}{\mathbf{A}^T \mathbf{XS} + \lambda \mathbf{S}^{-1/2} + \mu \mathbf{SW}} \tag{12}$$

To take care of ASC constraint on abundance fractions matrix that is given in Eq. (3), FCLS method proposed in (Heinz and Chang 2001) has been used in this paper. In FCLS method, new observation and signature matrices should be defined using the following equations.

$$\tilde{\mathbf{X}} = \begin{bmatrix} \mathbf{X} \\ \delta\mathbf{1} \end{bmatrix}, \tilde{\mathbf{A}} = \begin{bmatrix} \mathbf{A} \\ \delta\mathbf{1} \end{bmatrix} \tag{13}$$

where $\delta$ controls the impact of ASC constraint and $\mathbf{1}$ is the all one vector.

The initial values of $\mathbf{A}$ and $\mathbf{S}$ can be obtained using random initialization or VCA algorithm. Algorithm 1 summarizes the proposed algorithm. In this algorithm, $Th$ denotes the threshold of error between the original and reconstructed matrices and $T_{\max}$ is the maximum number of iterations.

Algorithm 1. Sparse GNMF Algorithm

| |
|---|
| Initialize $\mathbf{A}$ and $\mathbf{S}$ using VCA algorithm and FCLS method |
| $t = 1$ |
| Do |
|     $t = t + 1$ |
|     Update $\mathbf{A}$ using Eq. (11) |
|     Augment $\mathbf{A}$ and $\mathbf{S}$ using Eq. (13) |
|     Update $\mathbf{S}$ using Eq. (12) |
| Until $\|\mathbf{X} - \mathbf{AS}\| < Th$ or $t > T_{\max}$ |

Some of the parameters in the proposed algorithm are selected based on previously published papers and some of them are selected using trial and error method. However one can optimize the algorithm to find the best possible values of the parameters.

## 3. Evaluation Results

### 3.1. Performance Metrics

To evaluate the proposed algorithm, synthetic and real datasets have been used in this paper. Two quantitative measures are used to quantify the results and compare the performance of the proposed method with two other well established methods. Spectral angle distance (SAD) and abundance angle distance (AAD) are the measures to compute the difference between estimated spectral signatures ($\hat{m}_p$) and abundance fractions ($\hat{a}_i$) with the original ones ($m_p$ and $a_i$) respectively. These measures are given in the following equations:

$$\text{SAD}_{m_p} = \cos^{-1}\left(\frac{m_p^T \hat{m}_p}{\|m_p\|\|\hat{m}_p\|}\right) \tag{14}$$

$$\text{AAD}_{a_i} = \cos^{-1}\left(\frac{a_i^T \hat{a}_i}{\|a_i\|\|\hat{a}_i\|}\right) \tag{15}$$

These measures are defined for each endmember and each abundance fractions vector. To have an overall measure for all endmembers and abundance fractions, RMS values of them can be calculated using the following equations:

$$\text{rms}_{\text{SAD}} = \left(\frac{1}{P}\sum_{p=1}^{P}\left(\text{SAD}_{m_p}\right)^2\right)^{1/2} \tag{16}$$

$$\text{rms}_{\text{AAD}} = \left(\frac{1}{N}\sum_{i=1}^{N}\left(\text{AAD}_{a_i}\right)^2\right)^{1/2} \tag{17}$$

3.2. Synthetic dataset #1 using USGS spectral library

Usually, two kinds of datasets can be used to evaluate unmixing algorithms: synthetic and real datasets. Synthetic data has the advantage of having known set of endmember and abundance fractions. In this paper, the procedure described in (Miao and Qi 2007) is used. Synthetic image of size 64 by 64 has been generated without any pure pixels. To do so, first this image has been divided into 8 by 8 blocks. Then each block has been filled using six different materials that are selected from USGS spectral library version 06 (Clark et al. 2007) as illustrated in Fig. 3. Spatial low pass filter (LPF) is applied on the image to create linear mixture of the materials in pixels. The size of spatial LPF is 9 by 9 in this paper. At the end, zero mean Gaussian noise should be added to data to resemble measurement and sensor noise.

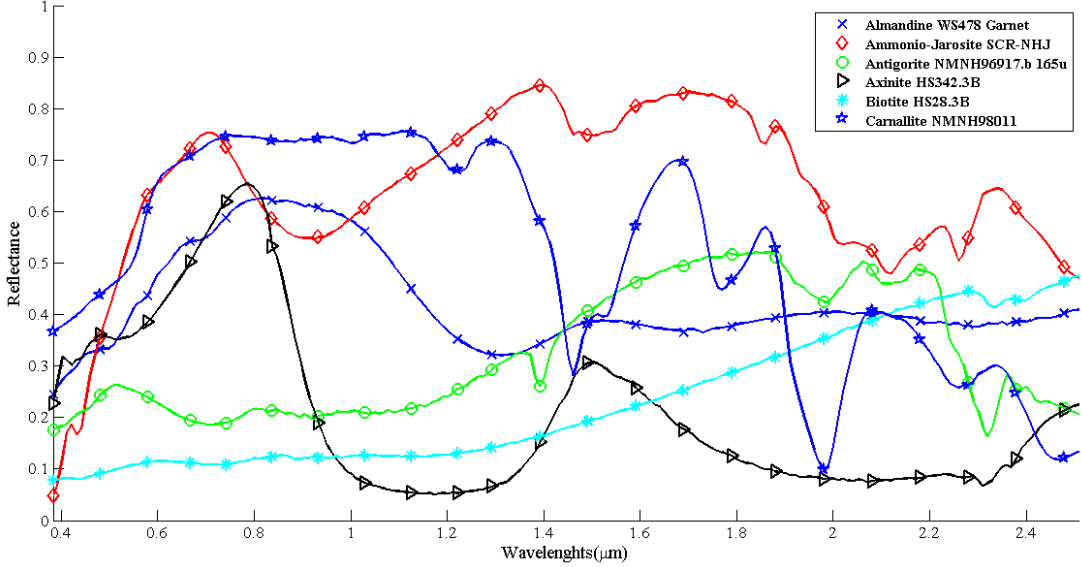

Fig. 3. Selected materials from USGS library to generate synthetic data

The proposed algorithm (sparse GNMF) has been applied to generated synthetic data with different levels of SNR ranging from 15 to 40dB. For this experiment the parameters are set as follows: $\lambda_0 = 0.05$, $\tau = 25$, $\mu = 0.1$, $\delta = 15$, $Th = 0.5 \times 10^{-3}$, $T_{max} = 3000$. The results in comparison with VCA (Nascimento and Dias 2005) and sparse NMF (Qian et al. 2011) methods are shown in Fig. 4 and Fig. 5 in terms of rmsSAD and rmsAAD respectively. Results show that the proposed method can give better results in terms of both evaluation metrics. Note that initialization of the algorithm by VCA does not yield deterministic results in each run, so we used monte carlo simulation to evaluate the algorithms and the presented plots are the mean value of the results for total number of runs.

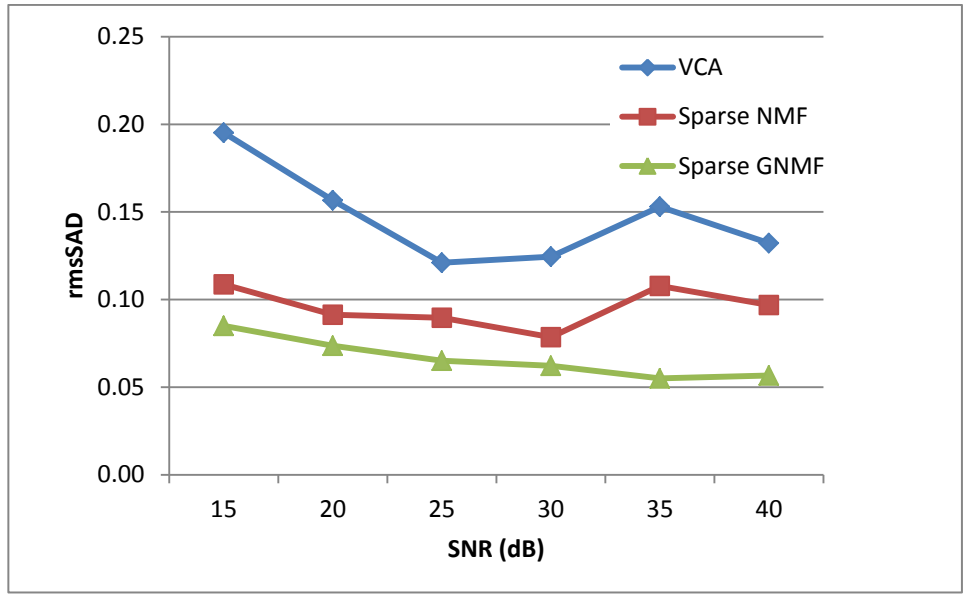

Fig. 4. Comparison of methods in terms of rmsSAD in radian vs. SNR

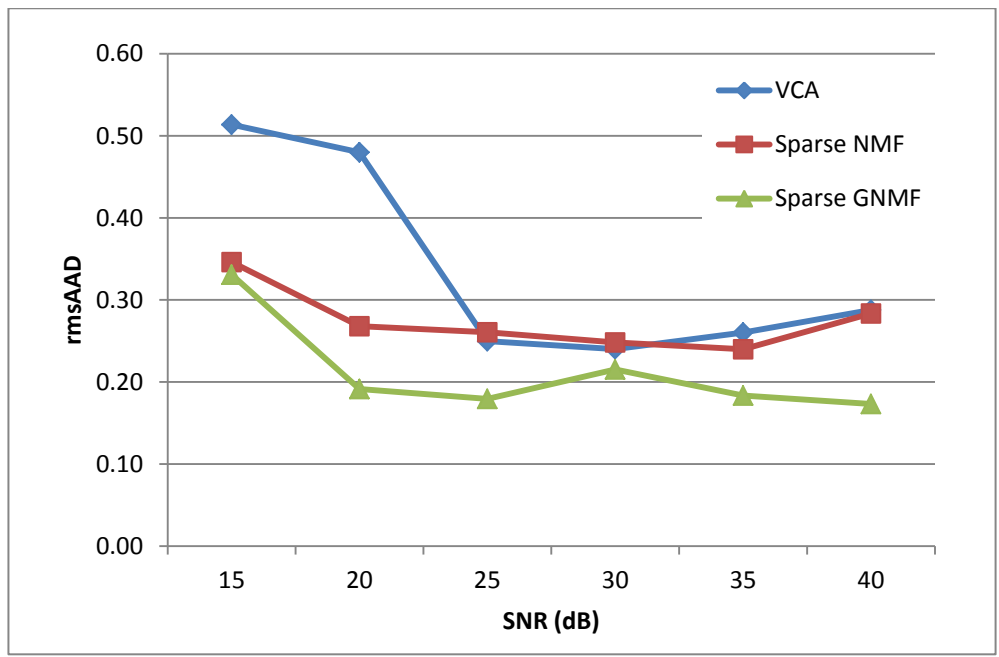

Fig. 5. Comparison of methods in terms of rmsAAD in radian vs. SNR

To show the convergence of the proposed algorithm, norm of error between original observation matrix and reconstructed one has been plotted against number of iterations in Fig. 6 for SNR=25dB. The maximum number of iterations in this plot is for the case of reaching stopping criteria as defined in Algorithm 1.

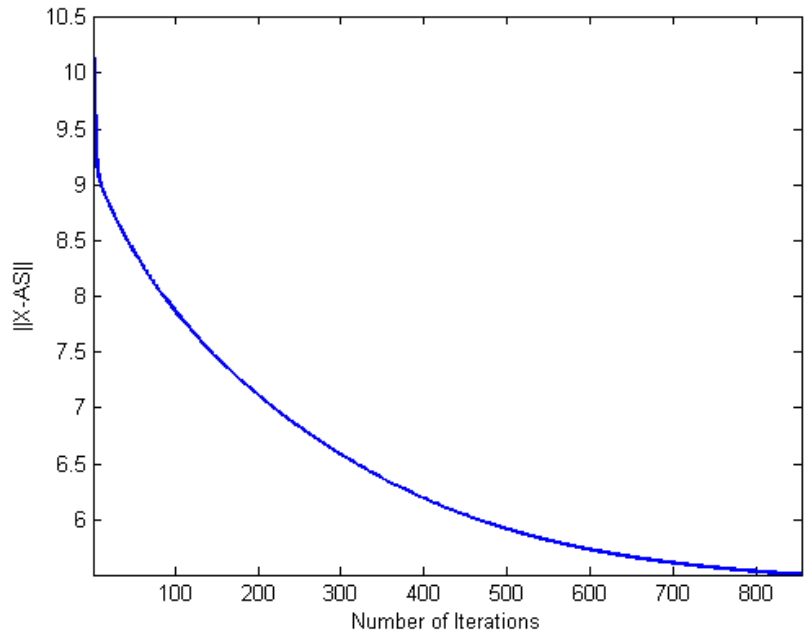

Fig. 6. ||X-AS|| vs. number of iterations in sparse GNMF method for SNR=25 dB in experiment I

3.3. Synthetic dataset #2 using USGS spectral library

In this experiment another synthetic dataset using different materials from USGS spectral library has been used to evaluate the proposed method. The purpose of this experiment is evaluating the performance of the proposed algorithm for spectral signatures that are more similar together. Selected materials for this test case are: Kaolinite, Gibbsite, Lepidolite, Montmorriolinte, Muscovite, Goethite, Hematite, and Limonite as shown in Fig. 7.

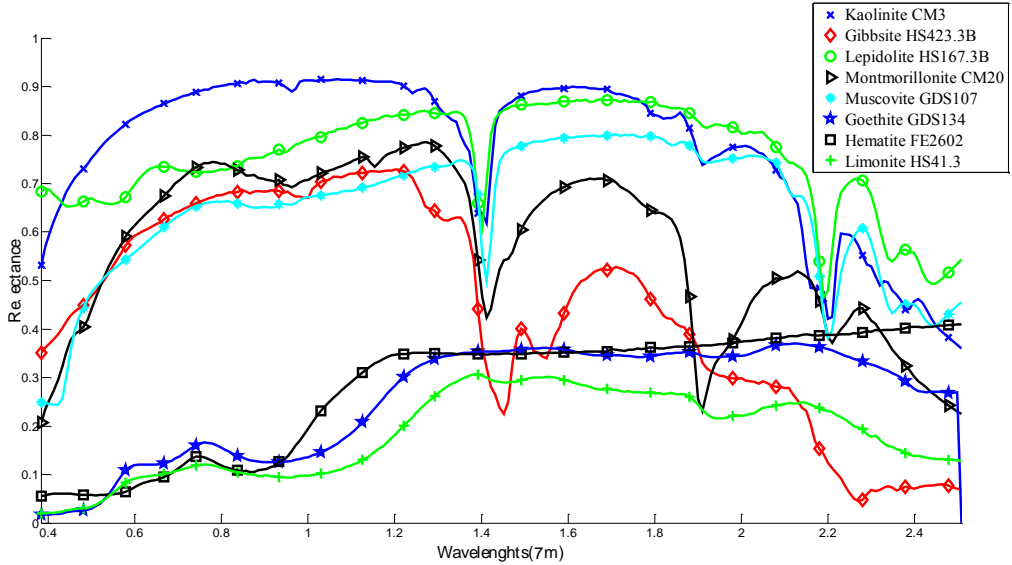

Fig. 7. Selected materials from USGS library to generate synthetic data for experiment II

In this experiment additive white Gaussian noise with SNR=25dB has been added to the synthetic data. Other parameters are set like the previous experiment in section 3.2. Fig. 8 demonstrated the estimated spectral signatures using the proposed method. As it can be seen from this figure estimated endmembers can follow the original ones efficiently. Also Fig. 9 shows the convergence of the proposed algorithm in this experiment.

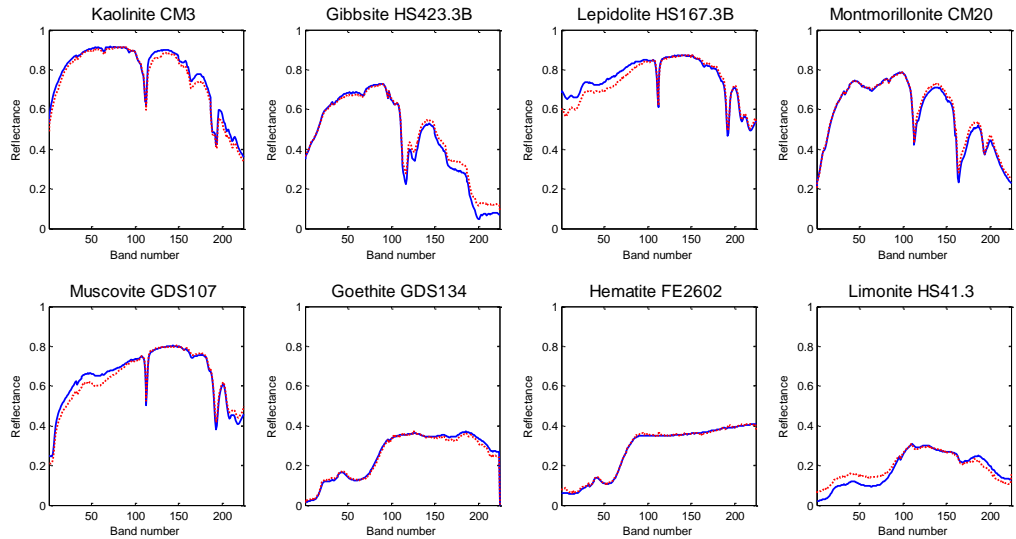

Fig. 8. Extracted spectral signatures using sparse GNMF (red dotted lines) in comparison with original ones (blue lines) for experiment II

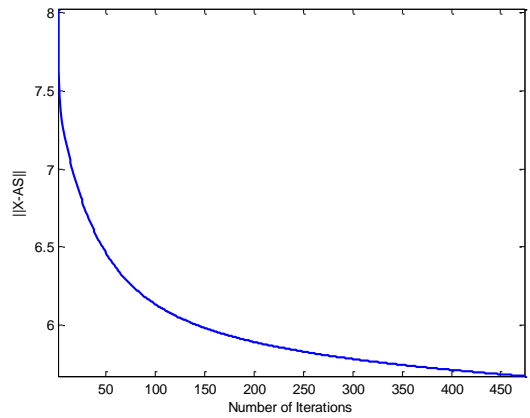

Fig. 9. ||X-AS|| vs. number of iterations in sparse GNMF method for SNR=25 dB in experiment II

## 3.4. Real dataset (Cuprite Nevada)

Another experiment has been done on real dataset. The Cuprite Nevada dataset collected by AVIRIS sensor is the most common public dataset[1] for the evaluation of hyperspectral unmixing algorithms that is also used in this paper. Band 30 of this dataset is illustrated in Fig. 10. The region is covered by mineral materials and data contains 250 by 191 pixels and 224 bands of spectral data. After removing low SNR and water absorption bands, 188 bands are remained and used in the experiment.

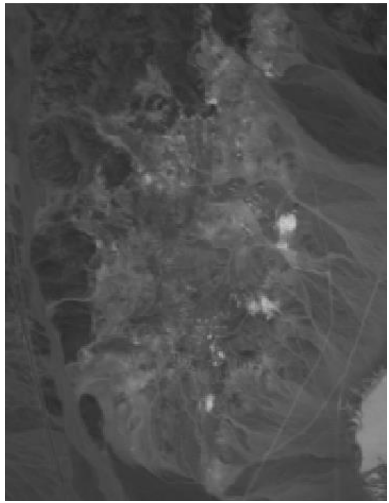

Fig. 10. Band 30 of the Cuprite Nevada dataset

---

[1] http://aviris.jpl.nasa.gov/data/free_data.html

The proposed algorithm is applied on this dataset. The estimated signatures are compared with reference spectral signatures from USGS AVIRIS convolved spectral library in terms of SAD. Table 2 summarizes the results in comparison with VCA and sparse NMF methods. In this table, the best results for each material and rmsSAD are emphasized in bold face font. As results show, the proposed algorithm can unmix the data more effectively in comparison with other methods. The outputs of the proposed method i.e. estimated spectral signatures and abundance maps are illustrated in Fig. 11 and Fig. 12 respectively. Note that due to observation noise in the process of hyperspectral data acquisition, the resulted spectra using spectral unmixing algorithms have some artifacts in comparison with USGS spectra that are measured in the laboratory. This is the reason behind the differences between some of estimated spectral signatures and original ones.

Table 2. Comparison results of Cuprite dataset in terms of SAD

| Material Name | VCA | Sparse NMF | Sparse GNMF |
| --- | --- | --- | --- |
| Alunite GDS82 Na82 | 0.1129 | 0.1814 | **0.1050** |
| Andradite WS487 Garnet | 0.0721 | 0.0716 | **0.0660** |
| Buddingtonite GDS85 D-206 | 0.1143 | **0.0773** | 0.1171 |
| Chalcedony CU91-6A | **0.1168** | 0.1344 | 0.1268 |
| Kaolin_Smect H89-FR-5 .3Kaol | 0.0699 | 0.0761 | **0.0600** |
| Kaolin_Smect KLF508 .85Kaol | 0.0856 | 0.0851 | **0.0652** |
| Dumortierite HS190.3B | 0.1529 | **0.0804** | 0.1021 |
| Montmorillonite+Illite CM37 | **0.0495** | 0.0585 | 0.0500 |
| Muscovite IL107 | **0.1176** | 0.1800 | 0.2113 |
| Nontronite NG-1.a | 0.0847 | 0.1243 | **0.0780** |
| Andradite WS474 | **0.0466** | 0.0691 | 0.1027 |
| Sphene HS189.3B | 0.2375 | 0.1379 | **0.0637** |
| rmsSAD | 0.1163 | 0.1143 | **0.1046** |

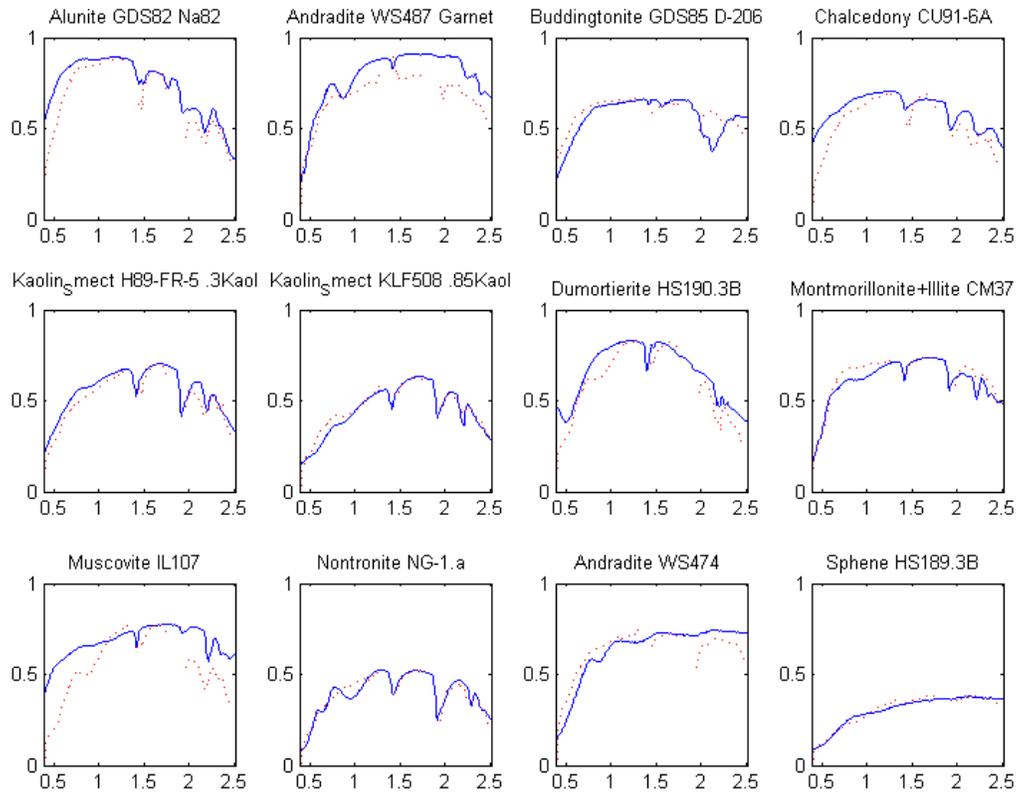

Fig. 11. Extracted spectral signatures (red dotted lines) in comparison with USGS library signatures (blue lines) of Cuprite dataset using sparse GNMF

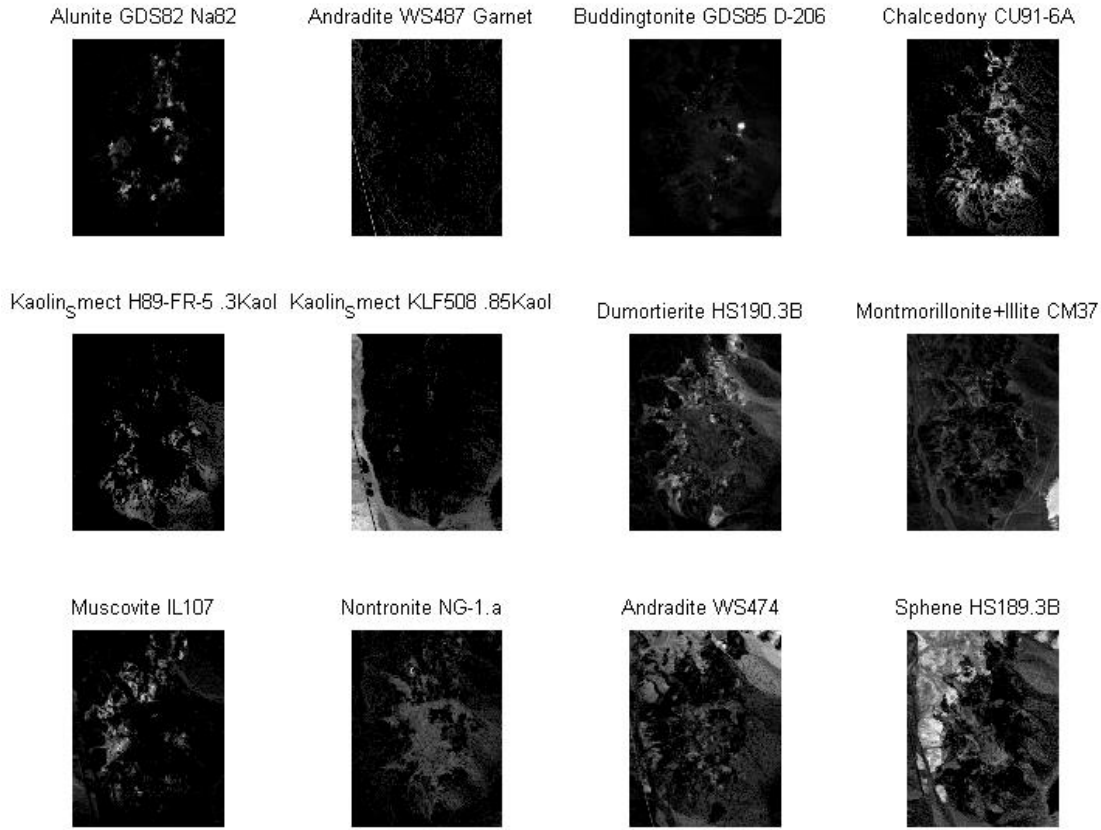

Fig. 12. Abundance fraction maps of Cuprite dataset using sparse GNMF

## 4. Conclusion

Mixed pixels processing is an active research area in hyperspectral remote sensing field. In this paper linear mixing model has been used to solve the spectral unmixing problem. This paper focuses on the category of nonnegative matrix factorization methods. NMF method does not have a unique solution, so it should be used in conjunction with other constraints to give promising results. In this paper two regularization terms have been added as constraints to the NMF problem. First constraint is sparseness constraint on abundance fractions. In this paper, the effect of this term is controlled using adaptive regularization parameter. The second constraint is graph regularization term. Graph regularized NMF will preserve the geometrical structure of data by defining a neighborhood graph on the observation data. KNN is used in this study to generate the neighborhood graph. To solve the optimization problem of the proposed sparse GNMF method, multiplicative update rules have been used. Performance evaluation of the proposed method has been done using SAD and AAD measures for spectral signatures and abundance fractions respectively. Synthetic and real datasets have been utilized to evaluate the proposed method. Results show that sparse GNMF can perform better in comparison with other well established methods. Specifically in AVIRIS Cuprite experiment, the results using the proposed method is about 10% improved in terms of rmsSAD.

References

Alizadeh, H., & Ghassemian, H. Hyperspectral data unmixing using constrained semi-NMF and PCA transform. In *20th Iranian Conference on Electrical Engineering (ICEE), 15-17 May 2012 2012* (pp. 1523-1527). doi:10.1109/IranianCEE.2012.6292600.
Bendoumi, M. A., & Mingyi, H. Unmixing approach for hyperspectral data resolution enhancement using high resolution multispectral image with unknown spectral response function. In *Industrial Electronics and Applications (ICIEA), 2013 8th IEEE Conference on, 19-21 June 2013 2013* (pp. 511-515). doi:10.1109/ICIEA.2013.6566422.
Cai, D., He, X., Han, J., & Huang, T. S. (2011). Graph Regularized Nonnegative Matrix Factorization for Data Representation. *IEEE Transactions on Pattern Analysis and Machine Intelligence, 33*(8), 1548-1560.
Chen, J., Richard, C., & Honeine, P. (2013). Nonlinear Unmixing of Hyperspectral Data Based on a Linear-Mixture/Nonlinear-Fluctuation Model. *IEEE Transactions on Signal Processing, 61*(2), 480-492, doi:10.1109/TSP.2012.2222390.
Cichocki, A., & Zdunek, R. (2006). Multilayer nonnegative matrix factorisation. *Electronics Letters, 42*(16), 947-948.
Cichocki, A., Zdunek, R., & Amari, S. I. New Algorithms for Non-Negative Matrix Factorization in Applications to Blind Source Separation. In *IEEE International Conference on Acoustics, Speech and Signal Processing, ICASSP 2006, 14-19 May 2006* (Vol. 5, pp. V-V). doi:10.1109/ICASSP.2006.1661352.
USGS digital spectral library splib06a http://speclab.cr.usgs.gov/spectral.lib06 (2007). U.S. Geological Survey. http://speclab.cr.usgs.gov/spectral.lib06.
Heinz, D. C., & Chang, C.-I. (2001). Fully constrained least squares linear spectral mixture analysis method for material quantification in hyperspectral imagery. *IEEE Transactions on Geoscience and Remote Sensing, 39*(3), 529-545.
Keshava, N. (2003). A Survey of Spectral Unmixing Algorithms. *Lincoln Lab Journal, 14*(1), 55-78.
Li, C., & Yin, J. (2013). A Multispectral Remote Sensing Data Spectral Unmixing Algorithm Based on Variational Bayesian ICA. *Journal of the Indian Society of Remote Sensing, 41*(2), 259-268, doi:10.1007/s12524-012-0245-0.
Ma, W., Bioucas-Dias, J., Gader, P., Chan, T., Gillis, N., Plaza, A., et al. (2014). Signal processing perspective on hyperspectral unmixing. *IEEE Signal Processing Magazine, 31*(1), 67-81.
Miao, L., & Qi, H. (2007). Endmember Extraction From Highly Mixed Data Using Minimum Volume Constrained Nonnegative Matrix Factorization. *IEEE Transactions on Geoscience and Remote Sensing, 45*(3), 765-777, doi:10.1109/tgrs.2006.888466.
Nascimento, J. M. P., & Dias, J. M. B. (2005). Vertex component analysis: a fast algorithm to unmix hyperspectral data. *IEEE Transactions on Geoscience and Remote Sensing, 43*(4), 898-910.
Pauca, V. P., Piper, J., & Plemmons, R. J. (2006). Nonnegative matrix factorization for spectral data analysis. *Linear Algebra and its Applications, 416*(1), 29-47, doi:10.1016/j.laa.2005.06.025.
Qian, Y., Jia, S., Zhou, J., & Robles-Kelly, A. (2011). Hyperspectral Unmixing via L1/2 Sparsity-Constrained Nonnegative Matrix Factorization. *IEEE Transactions on Geoscience and Remote Sensing, 49*(11), 4282-4297.
Rajabi, R., & Ghassemian, H. Graph Regularized Nonnegative Matrix Factorization for Hyperspectral Data Unmixing. In *7th Iranian Machine Vision and Image Processing (MVIP 2011), Tehran, Iran, 2011*
Rajabi, R., & Ghassemian, H. Hyperspectral data unmixing using GNMF method and sparseness constraint. In *IEEE International Geoscience and Remote Sensing Symposium (IGARSS 2013), , Melbourne, Australia, 2013*: IEEE
Remon, A., Sanchez, S., Bernabe, S., Quintana-Orti, E., & Plaza, A. (2013). Performance versus energy consumption of hyperspectral unmixing algorithms on multi-core platforms. *EURASIP Journal on Advances in Signal Processing, 2013*(1), 68.
Sanjeevi, S., & Barnsley, M. J. (2000). Spectral unmixing of compact airborne spectrographic imager (CASI) data for quantifying sub-pixel proportions of biophysical parameters in a coastal dune system. *Journal of the Indian Society of Remote Sensing, 28*(2-3), 187-204, doi:10.1007/bf02989903.

Villa, A., Chanussot, J., Benediktsson, J. A., & Jutten, C. (2011). Spectral Unmixing for the Classification of Hyperspectral Images at a Finer Spatial Resolution. *IEEE Journal of Selected Topics in Signal Processing, 5*(3), 521-533.

Yang, Z., Zhou, G., Xie, S., Ding, S., Yang, J.-M., & Zhang, J. (2011). Blind Spectral Unmixing Based on Sparse Nonnegative Matrix Factorization. *IEEE TRANSACTIONS ON IMAGE PROCESSING, 20*(4), 1112-1125, doi:10.1109/tip.2010.2081678.